%% file: main.tex
\documentclass{article} 
\usepackage{iclr2022_conference,times}

\input{math_commands.tex}

\usepackage{hyperref}
\usepackage{url}

\usepackage{dsfont}    
\usepackage{amsmath,amssymb}
\usepackage{stmaryrd}
\usepackage{amsthm}
\usepackage{multirow}
\usepackage{graphicx}
\usepackage{listings}
\usepackage{microtype}
\usepackage{graphicx}
\usepackage{wrapfig}
\usepackage{booktabs} %

\newcommand{\mean}[2]{\mathbb{E}_{#2} \! \left[ #1 \right]}

\title{Knowledge is reward: Learning optimal exploration by predictive reward cashing}

\author{Luca Ambrogioni \\
Donders Institute\\
Radboud University\\
Nijmegen, Netherlands \\
\texttt{l.ambrogioni@donders.ru.nl} 
}

\iclrfinalcopy 
\begin{document}

\maketitle

\begin{abstract}
There is a strong link between the general concept of intelligence and the ability to collect and use information. The theory of Bayes-adaptive exploration offers an attractive optimality framework for training machines to perform complex information gathering tasks. However, the computational complexity of the resulting optimal control problem has limited the diffusion of the theory to mainstream deep AI research. In this paper we exploit the inherent mathematical structure of Bayes-adaptive problems in order to dramatically simplify the problem by making the reward structure denser while simultaneously decoupling the learning of exploitation and exploration policies. The key to this simplification comes from the novel concept of cross-value (i.e. the value of being in an environment while acting optimally according to another), which we use to quantify the value of currently available information. This results in a new denser reward structure that "cashes in" all future rewards that can be predicted from the current information state. In a set of experiments we show that the approach makes it possible to learn challenging information gathering tasks without the use of shaping and heuristic bonuses in situations where the standard RL algorithms fail.

\end{abstract}

\section{Introduction}
\begin{figure}[ht]
    \centering
    \includegraphics[width=0.6\textwidth]{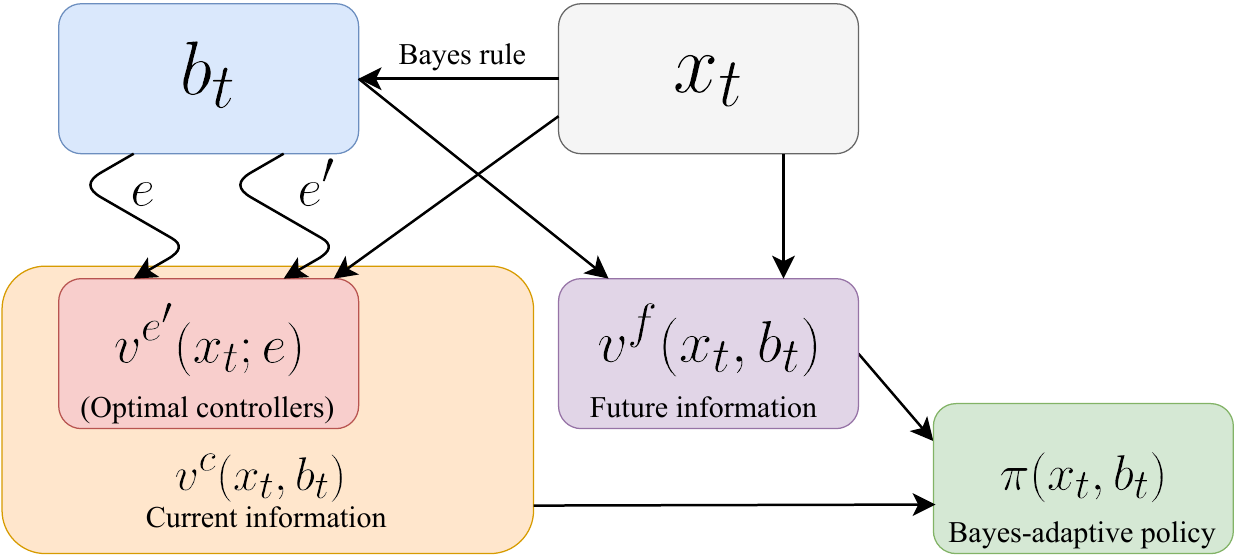}
   \caption{Diagram of a modular predictively cashed RL architecture. }
    \label{fig: diagram}
\end{figure}
We live in a world of information. In our daily lives, almost every novel task inevitably starts with a series of targeted information retrieval actions, whether performing an internet search or just asking a question to a friend. It is therefore imperative for the deployment of artificial intelligence in the real world to develop efficient algorithms capable of learning how to find and use information in complex environments. In the context of Bayesian reinforcement learning (RL), information and knowledge are formalized as the concept of belief state (also known as information state) \citep{ghavamzadeh2015}. One of the key results of Bayesian RL is that optimal exploration (also known as \emph{Bayes adaptive} exploration), and consequently optimal information retrieval, can be achieved by augmenting the state space with the belief state and solving a standard control problem on this augmented space \citep{ghavamzadeh2015}. In spite of its elegance, the very high dimensionality of belief-augmented problems, together with the intrinsic intractability of Bayesian inference, limits the range of applicability of this approach. Perhaps for this reason, most of the modern deep reinforcement learning literature adopts heuristic exploration methods such as $\epsilon$-greedy, Thompson sampling \citep{osband2015bootstrapped, osband2016deep, azizzadenesheli2018efficient}, noisy networks \citep{fortunato2017noisy}, random network exploration bonuses \citep{burda2018exploration} and curiosity-based approaches \citep{still2012information, frank2014curiosity, zhelo2018curiosity}. Recent developments in variational inference and deep reinforcement learning re-ignited the interest in approximate Bayes-adaptive exploration as a form of meta-reinforcement learning \citep{duan2016rl, zintgraf2019varibad, dorfman2020offline}. Nevertheless, learning targeted exploratory actions remains a very challenging task even with modern deep RL techniques, given the high dimensionality of the belief space and the fact that information gathering actions are only very indirectly related to the collection of reward. Fortunately the situation is less dire than it could superficially appear as belief-augmented RL problems inherit a considerable amount of structure from the underlying full information control problems and the nature of iterated Bayesian inference. In this paper we exploit this mathematical structure in order to simultaneously address two related sub-problems that make this setting hard: I) The sparsity of the reward structure \citep{zintgraf2021exploration} and II) the so called "chicken-and-egg problem" of exploration and exploitation \citep{liu2021decoupling}. This latter problem arises from the fact that the value of information gathering depends on its use to collect reward but a proper exploitation policy depends on the use of the available information. 
The other main problem limiting the use of Bayes-adaptive learning is the inherent computational complexity of non-conjugate iterated Bayesian inference. This paper does not deal with this problem and we recommend the reader to the extensive literature on approximate Bayesian inference \citep{zhang2018advances, blei2017variational, betancourt2017conceptual, sarkka2013bayesian}. Therefore, we limit our analysis to conjugate Bayesian models where inference can be performed using a simple update rule. However, all techniques introduced here can be straightforwardly applied to approximate inference settings, including modern approaches using variational autoencoders (VAEs) such as \citep{zintgraf2019varibad, dorfman2020offline}, possibly in combination with training bonuses methods such as in \citep{zintgraf2021exploration}.

\section{Related work}
The theoretical framework for optimal control on a belief-augmented state space was first developed in the control theory literature under the name of dual control theory \citep{feldbaum1960dual1, feldbaum1960dual2, feldbaum1961dual3, filatov2000survey}. The high computational complexity of the approach, stemming from the very high dimensionality of the state space, limited the diffusion of the theory. However, the approach was re-introduced in the context of model-based Bayesian reinforcement learning \citep{rieder1975bayesian, ross2007bayes} where the domain of applicability was greatly increased by the introduction of Monte Carlo tree search methods \citep{guez2012efficient, guez2013scalable, asmuth2012learning, katt2017learning}. The use of belief-augmentation in model-free RL methods is very limited. To the best of our knowledge, the first use of temporal difference (TD) learning to solve dual control problems is in \cite{santamar1997new}. Until recently, the deep reinforcement learning literature ignored the concept of Bayes-adaptive exploration, belief-augmentation and dual control. However, important progresses have been made in the related field of model-free meta-RL \citep{wang2016learning, duan2016rl, gupta2018meta}. For example, $\text{RL}^2$ involves the training of a recurrent meta-network in order to learn optimal exploration in novel environments \citep{duan2016rl}. This can be seen as a model-free form of Bayes-optimal learning since the information encoded in a Bayesian belief state is fully contained in the sequence of past transitions that are fed into the recurrent network. However, this results in redundant encoding as it ignores the exchangability of the observations, which in turn can make an already difficult learning problem into something substantially harder. The recently proposed variBAD method is the first work using modern deep learning techniques in an explicitly belief-augmented setting \cite{zintgraf2019varibad}. This work uses VAEs for performing the approximate inference and hence obtaining the belief-state and then exploits a policy gradient method in order to obtain an approximate Bayes-adaptive policy. Unfortunately, the extremely sparse reward structure of most useful optimal exploration tasks makes methods such as variBAD very challenging to train. The HyperX method addresses this problem with a shaping approach by adding a series of reward
bonuses to promote (meta-)exploration of the belief-augmented state space during training \citep{zintgraf2021exploration}. While our paper deals with the same underlying problem, our solution involves the reformulation of the reward structure into an equivalent one without the use of any heuristic shaping bonus. In this sense, our approach and HyperX are orthogonal and potentially complementary and can be used together to solve more challenging targeted exploration tasks. Similarly to our work, the recently introduced DREAM method decouples the learning of the exploitation and exploration modules so as to ameliorate the "chicken-and-egg" problem that arises from their interdependence \citep{liu2021decoupling}. However, the approach used in DREAM is radically different from ours, as it exploits the environment ID and uses an information bottleneck loss to extract task relevant information. Our predictively cashed reward bears some superficial resemblance to curiosity-based approaches such as \citep{schmidhuber1991possibility, oudeyer2007intrinsic, frank2014curiosity, stadie2015incentivizing, bellemare2016unifying, tang2016exploration, ostrovski2017count, zhao2019curiosity} since it directly rewards the discovery of new information. However, while curiosity rewards information in itself, the predictively cashed reward only values the difference in predicted reward that the new information allows to collect. 

\section{Preliminaries}
Markov decision processes (MDPs) are a mathematical formalization of sequential decision making in stochastic environments. A MDP is defined by: 1) a family of stochastic processes specified by a set of transition probabilities $p\!\left(x_{t+1}\mid x_{t}, a_{t}\right)$ between states $x_t$ that are conditional on a sequence of action variables $a_t$ and 2) a family of reward probabilities $p\!\left(r_t \mid x_t, a_t \right)$. In a RL problem, transition and reward probabilities  are not known in advance. Without loss of generality, we can formalize this uncertainty by making the transition and reward probabilities dependent on a set of "environment" parameters $e$:
\begin{align} \label{eq: transition}
    &p\left(x_{t+1}\mid x_{t}, a_{t}\right) = T\left(x_{t+1}\mid x_{t}, a_{t} , e\right)~,\\ \label{eq: reward}
    &p \left(r_t \mid x_t, a_t \right) = R \left(r_t \mid x_t, a_t, e \right)~.
\end{align}
This model describes a family of Markov decision processes parameterized by the environment variable $e$, which is not directly accessible to the agent and can only be inferred from the observed transitions/rewards. A policy $\pi(a_t \mid \{\ x_\tau, a_\tau, r_\tau\}_{\tau=0}^{t-1})$ is a distribution over actions conditional on a previous sequence of state transitions. Every policy $\pi$ defines a $\pi$-controlled process where the actions are sampled according to the policy and state transitions and rewards are sampled accordingly. We denote the expectation under a $\pi$-controlled process in an environment $e$ as $\mean{\cdot}{\pi \mid e}$. The discounted expected value (often simply referred to as just the value) of a policy $\pi$ in an environment $e$ is defined as:
\begin{equation}\label{eq: optimal value of e}
    v^\pi(x_t; e) = \mean{\sum_{\tau = t}^{\infty} \gamma^{\tau - t} r_t}{\pi \mid e}~,
\end{equation}
where $\gamma \in (0, 1)$ is a discounting factor and the sequence of states and rewards is sampled from the $\pi$-controlled stochastic process. The optimal value $v^e(x_t; e)$ is defined as the global optimum of Eq.~\ref{eq: optimal value of e} with respect to the policy (the reason for this unusual notation will be clear later on). Now assume that environment $e$ is sampled from a prior distribution $p(e; b_0)$, parameterized by the initial "belief state" $b_0$. This initial belief state summarizes all the information available to the agent at time zero. After each transition 
$
x_t, a_t, r_t \rightarrow x_{t+1}
$
the belief state is updated using Bayes rule:
\begin{equation}
    p(e; b_{t+1}) = p(e \mid x_t, x_{t+1}, a_t, r_t) \propto R \left(r_t \mid x_t, a_t, e \right) T\left(x_{t+1}\mid x_{t}, a_{t} , e\right) p(e; b_t)~.
\end{equation}
Here we assumed that the posterior distribution can be parameterized by the belief variables $b_{t+1}$ given any possible transition. This is only approximately possible in non-conjugate cases, unless we allow for the use of  infinite dimensional belief states (in that case the belief state can be chosen to be the posterior density itself). The appropriate objective function for learning optimal exploration is then simply the average of the environment specific values under the belief state: 
\begin{equation}\label{eq: bayes value}
    v^\pi(x_t, b_t) = \mean{v^\pi_t(x_t; e)}{e \sim b_t} = \mean{\mean{\sum_{\tau = t}^{\infty} \gamma^{\tau - t} r_\tau}{\pi \mid e}}{e \sim b_t}~.
\end{equation}
A Bayes-adaptive policy is a global maximizer of Eq.~\ref{eq: bayes value}. We denote the optimal belief-augmented value as $v^*(x_t,b_t)$. In order to maximize the expected value, the agent needs to perform exploratory actions to update its belief state and then use the information to collect reward. Roughly speaking, these dynamics can be separated into an initial exploration phase and a subsequent exploitation phase. The optimal policy is non-Markov with respect to the state variable since the information at time $t$ concerning the environment $e$ depends on past transitions. However, the policy is Markov if state $x_t$ is augmented by the belief state $b_t$. The computational challenge of belief-augmented reinforcement learning led the community towards more tractable alternatives. A simple option is Thompson sampling, where the belief-augmented policy is approximated by the optimal policy of an environment $e'$ sampled from the belief state:
$
\pi(x_t, b_t) \approx \pi(x_t; e'), \text{ with } e' \sim b_t~.
$
However, a posterior sampling agent cannot learn how to perform actions that are not optimal in any known environment in the full information regime. In other words, the agent cannot learn to perform actions such as asking questions, consulting a map or performing an internet search as these actions are solely aimed at acquiring new information. Belief-augmentation allows us to solve the optimal exploration problem, also known as the exploration/exploitation dilemma, using standard reinforcement learning techniques. For example, given a belief-augmented transition 
$
x_t, b_t, a_t, r_t \rightarrow x_{t+1}, b_{t+1}~
$
under a belief-augmented policy $\pi(x_t, b_t)$, we can learn the expected value of the policy using the standard TD learning (tabular) update rule
\begin{equation} \label{eq: belief-augmented update}
    v^\pi(x_t, b_t) \shortleftarrow v^\pi(x_t, b_t) + \eta \left( r_t + \gamma v_{t+1}^\pi(x_{t+1}, b_{t+1}) - v^\pi(x_t, b_t) \right)~,
\end{equation}
where $\eta$ is a learning rate. The key difference when compared with standard TD learning is that the value is now a function of the belief state. This allows to assign high values to belief states where the agent has precise information concerning how to collect reward. For example, the value of being on a Caribbean island and knowing that it hides a buried pirate treasure is higher than the value of being in the same island without that knowledge. This is true both because the belief correlates with the actual state of the world and because it affects the optimal policy (e.g. digging vs sunbathing). 

\section{Cross-values and the value of current and future information}
We are now in the position to introduce our main contribution. Here we will show that a simplified version of the posterior sampling policy can be used to quantify the value of the information currently held by the agent. The expected value of a posterior sampling policy is not easy to quantify as each action leads to new belief updates. However, we can evaluate the value of a simplified sampling policy where the environment is sampled once from the belief and then the policy acts optimally until the end of the episode according to the sampled environment. In this case, the expected value is
\begin{equation} \label{eq: avg cross value}
    \mean{v^{e'} (x_t; e)}{e' \sim b_t}~. 
\end{equation}
We refer to the quantity $v^{e'} (x_t; e)$ as the \emph{cross-value}, which is defined as the expected reward collected in environment $e$ under the optimal policy of environment $e'$. The cross-value plays a crucial role in the present work as it will allow us to quantify the penalty that the agent has to pay when acting under "wrong beliefs". Our starting point is Eq.~\ref{eq: bayes value}, which expresses the optimal belief-augmented value as an average of environment-specific values: 
\begin{align} \nonumber 
    v^*(x_t,b_t) &=  \mean{\mean{\sum_{\tau=t}^{\infty} \gamma^{\tau - t} r_\tau}{{\pi \mid b_t}; e}}{e \sim b_t} \\
    &= \mean{v^*(x_t,b_t \mid e)}{e \sim b_t}~, \label{eq: optimal ba value}
\end{align}
where $v^*(x_t,b_t \mid e)$ is the value of the optimal belief-augmented policy when the agent is in environment $e$. The value $v^*(x_t,b_t, a_t \mid e)$ still depends on the belief state through the belief-augmented policy. However, an interesting belief independent approximation can be obtained by replacing the belief-augmented value $v^*(x_t,b_t, a_t \mid e)$ with the posterior sampling value given in Eq.~\ref{eq: avg cross value}:
\begin{align} 
    v^c(x_t,b_t) &= \mean{\mean{v^{e'} (x_t; e)}{e' \sim b_t}}{e \sim b_t} = \mean{v^{e'} (x_t; e)}{e,e' \underset{\text{iid}}{\sim} b_t}~.
\end{align}
We refer to $v^c$ as the \emph{value of current information} as it can it interpreted as the value of a policy that stops acquiring information and starts acting on a fixed environment sampled according to its current belief. This quantity has some important and intuitive properties. If we denote a deterministic belief state as $\delta_e$ (dirac measure), we have that $v^c(x_t,\delta_e) = v^e(x_t; e)$. In words, the value of future information is equal to the optimal value in the full information regime. More generally, the value of current information tends to increase as the entropy of the belief state decreases since the cross-values tend to converge to the optimal values. From the optimality of the value in Eq.~\ref{eq: optimal ba value} it follows that the value of current information is a lower bound on the optimal belief-augmented value:
\begin{equation}
    v^c(x_t,b_t) \leq v^*(x_t,b_t)~.
\end{equation}
Therefore, we can express the optimal belief-augmented value as a sum of the value of current information and a positive-valued term:
\begin{equation} \label{eq: value decomposition}
    v^*(x_t,b_t) = v^c(x_t,b_t) + v^f(x_t,b_t)~.
\end{equation}
We refer to $v^f(x_t,b_t, a_t)$ as the value of future information. Since the value of current information can be computed in a non-augmented state space, the belief-augmented problem reduces to learning the value of future information. 

\paragraph{Predictive reward cashing.}
The central result of this paper is that, once we have an expression for the cross-values, the full belief-augmented problem can be reduced to a simpler problem where information acquisition is directly rewarded. We can achieve this by plugging the decomposition of the value in Eq.~\ref{eq: value decomposition} into the belief-augmented update rule in Eq.~\ref{eq: belief-augmented update}. This results in a similar update rule for the value of future information:
\begin{equation} \label{eq: v update rule}
    v^f(x_t, b_t)  \shortleftarrow v^f(x_t, b_t) + \eta \left(\lambda_t + \gamma v_{t+1}^f(x_{t+1}, b_{t+1}) -  v^f(x_t, b_t)\right)~,
\end{equation}
where $\lambda_t$ is the \emph{predictively cashed} reward
\begin{align}\label{eq: modified reward}
    \lambda_t =r_t + \gamma v^c(x_{t+1}, b_{t+1}) - v^c(x_{t}, b_{t})~. 
\end{align}
This can been seen by plugging Eq.~\ref{eq: value decomposition} into the TD target:
\begin{align} \nonumber
    &r_t + \gamma v^\pi_{t+1}(x_{t+1}, b_{t+1}) -  v^\pi(x_t, b_t)  \\ \nonumber
    &= r_t + \gamma \left(v^c(x_{t+1}, b_{t+1}) + v^f(x_{t+1}, b_{t+1})\right) -  \left(v^c(x_t, b_t) + v^f(x_t, b_t)\right) \\ \nonumber
    &=  \lambda_t + \gamma v_{t+1}^f(x_{t+1}, b_{t+1}) -  v^f(x_t, b_t)~.
\end{align}

This quantity can be interpreted as the difference between the future expected reward collected in the past and present belief states if the agent were to stop updating its belief and act optimally according to an environment sampled from the belief. This can be seen by evaluating its expected value:
\begin{align} \nonumber
    \mean{\lambda_t}{r_t, x_{t+1}} &= \mean{r_t}{r_t} + \gamma v^c(x_{t+1}, b_{t+1}) - v^c(x_{t}, b_{t})\\ \nonumber
    &= \mean{ \mean{r_t}{r_t \mid e} + \gamma \mean{v^{e'}(x_{t+1}; e)}{x_{t+1}\mid \pi(e)}}{e,e'} - v^c(x_{t}, b_{t}) \\ \nonumber
    & \approx \mean{ \mean{r_t}{r_t \mid e} + \gamma \mean{v^{e'}(x_{t+1}; e)}{x_{t+1}\mid \pi^{e'}(e)}}{e,e'} - v^c(x_{t}, b_{t}) \\ 
    & = v^c(x_{t+1}, b_{t}) - v^c(x_{t}, b_{t}) \label{eq: approximate PCR}~,
\end{align}
where the approximation comes from the fact that we took the expectation with respect to the optimal policy under the environment $e'$ instead of the Bayes-adaptive policy. Therefore, the predictively cashed reward allows to immediately "cash in" future reward coming from an increase of available information. In practice, this transformation converts the very sparse reward structure of a belief-augmented problem into a much denser reward structure where reward is delivered directly as soon as relevant information is acquired. 

A key feature of the value of future information is that it can be bounded by full information quantities (i.e. quantities that can be computed without solving the belief-augmented problem). This is important as it means that we can bound function approximations of the value of future information without need for training, thereby ensuring the proper convergence of the belief-augmented values to the environment-specific optimal values as soon as the agent acquires information. The starting point is the inequality:
\begin{equation}
    v^*(x_t,b_t) = v^c(x_t,b_t) + v^f(x_t,b_t) \leq \mean{v^e(x_t; e)}{e \sim b_t}~.
\end{equation}
This result is obvious, it simply says that on average an optimal agent with full information collects at least as much reward as a belief-augmented agent with potentially incomplete knowledge of its environment. Therefore, $0 \leq v^f(x_t,b_t) \leq \mathcal{B}(x_t, b_t)~$, where the upper bound is given by
\begin{equation}\label{eq: bound}
    \mathcal{B}(x_t, b_t) = \mean{v^e(x_t; e) - \mean{v^{e'}(x_t; e)}{e' \sim b_t}}{e \sim b_t}~.
\end{equation}
This is the difference between the average expected future reward that can be acquired under the current information state and the average optimal expected reward. 

\section{Practical algorithms}
The predictively cashed reward (\ref{eq: modified reward}) can rarely be computed analytically even when the cross-value function is available. However, it is straightforward to obtain an unbiased stochastic estimator by replacing the exact expectation with an average over a finite number of samples:
$
    \lambda_t^{(M, N)} = r_t + \frac{1}{M N} \sum_{m=1}^M \sum_{n=1}^N v^{e_n} (x_t; e_m)
$,
with all the $e_k$ sampled independently from the belief state. We can therefore use this stochastic estimator as reward during training in place of the true $\lambda_t$ without affecting the asymptotic convergence. This is equivalent to switching from expected rewards to sampled rewards in conventional RL algorithms. In practice, we suggest to use $M = 1$. This generally provides a strong reward signal even when $N$ is small since the cross-value usually decreases sharply as the difference between the two environments increases, delivering strong reward signals when relevant information is acquired. All standard value-based RL algorithms can be used to learn the value of future information from this modified reward structure. The main difference between the predictively cashed training and standard training is that the policy should maximize the total value $v^*(x_t,b_t) = v^c(x_t,b_t) + v^f(x_t,b_t)$ instead of just the learned value $v^f(x_t,b_t)$. In practice, $v^c(x_t,b_t)$ has to be estimated from the cross-values using a finite number of samples. This leads to a stochasticity in the policy similar to standard Thompson sampling, which can potentially facilitate (meta-)exploration during training. Convergence of the tabular algorithm follows from the standard convergence results of RL. \textbf{Tabular approaches in finite deterministic inference settings}:
For the sake of simplicity, so far we presented the approach by its tabular update rule. Unfortunately, tabular algorithms are rarely feasible in a belief-augmented setting as there usually are infinitely many belief states even for very simple inference problems. However, there is an interesting class of problems where the full belief-augmented state space has a finite number of states. This happens when there is a finite number of possible environments and inference is deterministic, meaning that the belief variable can only transition from its prior value to a full information state. For a set of $S$ possible environments, the belief state can be parameterized by an array of posterior probabilities $b_t = (\rho^{(1)}_t,...,\rho^{(S)}_t)$. When inference is deterministic, the belief state can either stay on its prior value or "collapse" on a one-hot-encoded array representing a full information state. This implies that the belief space has $S + 1$ different states and, assuming that the environments have $H$ many states, the belief-augmented value can be represented by a table of $H$ numbers. This is true because we know that the value of future information is zero for full information states. Thus, the only non-zero values are in the starting prior belief state. While deterministic inference can sound very restrictive, it can effectively represent many different experimental behavioral and cognitive tasks in animals and humans where cues offer an unambiguous disambiguation of the latent state \citep{friston2016active}. \textbf{Function approximations with approximate value bound:}
In most realistic scenarios the belief-augmented state space has an exponentially large or infinite number of states and tabular algorithms become unfeasible. In these situations we can use standard function approximation techniques to estimate $v^f(x_t,b_t)$. We can also obtain a functional approximation of the value of future information that automatically satisfies the bound:
\begin{equation}\label{eq: function approximation with bound}
    v^f(x_t,b_t) = \mathcal{B}(x_t, b_t)w(x_t, b_t)~,
\end{equation}
where $w(x_t, b_t) \in (0,1)$ is the output of a parameterized function approximation such as a deep neural architecture. In practice, the bound $\mathcal{B}(x_t, b_t)$ can be estimated using a finite number of samples. This approximation ensures that the policy switches from exploratory to the optimal exploitative policy automatically as soon as enough relevant information has been collected. This is because the belief-augmented value is the sum of the value of current information, which converges to the optimal full-information value as the belief converges to a deterministic measure, and a term $v^f \leq \mathcal{B}$ that is guaranteed to vanish under this limit. Additionally and perhaps more importantly, the use of the bound in a functional approximation means that the agent does not need to learn how the predictively cashed reward coming from the same state is down-scaled when information is acquired. This is crucially important as the apparent inconsistency of the reward schedule makes the learning problem particularly hard.

\section{Learning the cross-values}
So far we assumed the cross-values to already be available before training. Here we will present some methods to either pre-compute the cross-values or to jointly train them together with the value of future information. When the expected reward and transition distributions are known, the optimal expected values can be obtained by solving the Bellman equation. It is straightforward to extend the approach in order to compute the cross-values, as described in Appendix A. Importantly, the resulting recursive system of equations can be efficiently parallelized on GPUs in a manner similar to convolutional neural networks. This allows to efficiently compute a batch of cross-values in order to estimate the expectations in Eq.~\ref{eq: modified reward} and Eq.~\ref{eq: bound}. In the stationary infinite horizon case, the time dependent system of equations is replaced by steady state value iterations as usual. When the reward and transition distributions are not available the cross-values can be learned using standard on-policy RL methods from sampled transitions. 
If the ground truth environment $e$ of each transition is available, the cross-value with respect to an arbitrary environment $e'$ can be updated using the tabular rule
\begin{equation} \label{eq: cross update}
    v^{e'}(x_t; e) \shortleftarrow v^{e'}(x_t; e) + \eta \left( r_t + \gamma v_{t+1}^{e'}(x_{t+1}; e) - v^{e'}(x_t; e) \right)~,
\end{equation}
where the action $a_t$ is chosen to maximize the current estimate of the optimal value $v^{e'}(x_t; e')$ through on-policy techniques such as policy gradients. In practice, the environment $e'$ can be sampled before each action from the belief state $b_t$, resulting in a Thompson sampling policy. 
\paragraph{Belief horizon sampling.}
As we have shown, it is straightforward to estimate the cross-value function when the ground truth environment $e$ is available at training time. However, in the most interesting settings the true environment can only be inferred through the agent's observations. In this situation, we propose the use of a method we call \emph{belief horizon sampling} where the "ground truth" environment is sampled from the terminal belief state of the episode $b_T$, with $T$ the last time point of the episode. This is equivalent to training with the actual ground truth for $T$ tending to infinity, as long as the $\pi$-controlled process satisfies the convergence conditions of the posterior to the maximum likelihood estimate. However, this approach also works well when the posterior is far from convergence since, loosely speaking, the sequence $\{\ x_\tau, a_\tau, r_\tau\}_{\tau=0}^{T}$ has high probability under all the conditional distributions $p(\{\ x_\tau, a_\tau, r_\tau\}_{\tau=0}^{T} \mid e')$ with $e'$ in the high probability region of the posterior $p(e \mid \{\ x_\tau, a_\tau, r_\tau\}_{\tau=0}^{T})$. In appendix B we show that this converges to the correct cross-values in the special case of deterministic inference. 

\section{Modular Bayes-adaptive deep reinforcement learning}
From the point of view of deep RL, the simultaneous learning of cross-values through belief horizon sampling and of the value of future information using predictively cashed rewards results in a modular architecture with components trained with separate loss functions from the same experienced transitions, as visualized in Fig.~\ref{fig: diagram}. On one hand, the cross-values are estimated by a deep net that takes $e$, $e'$ and $x$ and outputs the cross-value (red module in the figure). On the other hand, another deep RL architecture is used to approximate the value of future information from predictively cashed rewards estimated using the cross-values (violet module). When inference cannot be performed in closed-form, an additional inference network needs to be trained using techniques such as variational inference \cite{zintgraf2019varibad} (blue module). Finally, the policy network is trained to maximize the estimated belief-augmented value (green module).
    
\section{Experiments}
In this section we investigate the performance of the approach in a series of problems with closed-form Bayesian updates. In particular, we focus on problems where there is either an action or a state whose purpose is to reveal the location of reward. 

\paragraph{Disambiguation from a finite set of environments.}

\begin{figure}[ht]
    \centering
    \includegraphics[width=1\textwidth]{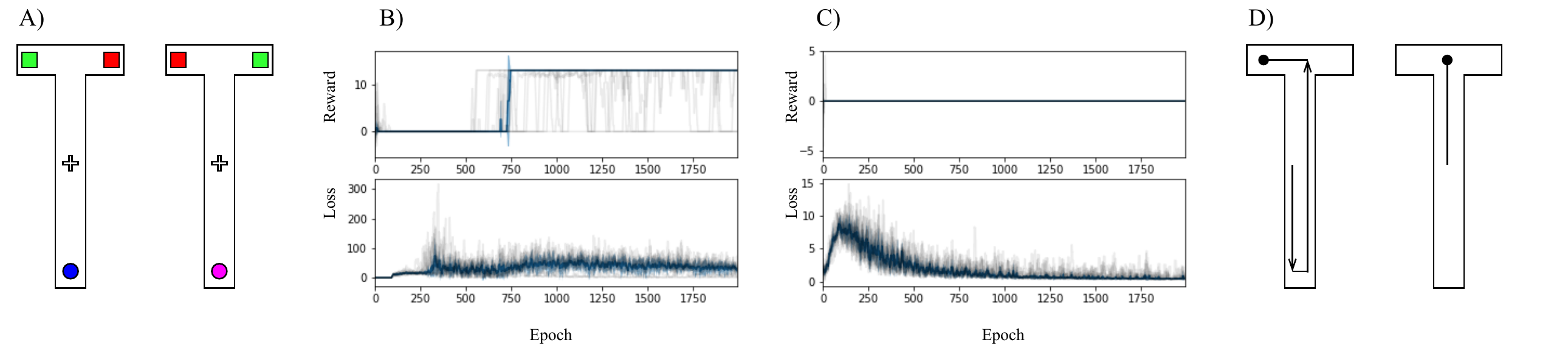}
   \caption{Maze disambiguation experiment. A) The two possible environments. The green square represents reward (+1) while the red square represents punishment (-7). The blue/purple circles are disambiguation cues that indicate the location of the reward. The optimal course of action is to visit the cue and then go towards the reward. B) Results of tabular PCR q-learning. The black line denotes the median collected reward (q-loss for bottom panel) over 15 repetitions as a function of epoch. The shaded blue area shows the median absolute deviation from the median. Individual transparent lines are the result of individual repetitions. C) Results of the standard tabular q-learning algorithm. D) Visualization of the (median) trained policy of PCR-TD (left diagram) and TD (right diagram) in the maze with left reward. The terminal black dot signifies that the agent stays at the location until the end of the episode. The PCR-TD policy is optimal. The TD policy learns to approach the reward region and then stops as it cannot disambiguate reward from punishment, resulting in zero total collected reward.}
    \label{fig: t maze result}
\end{figure}

We start with a single experimental set up inspired by animal neuroscience research \citep{friston2016active, wauthier2021learning}. The environment is a T-shaped maze with the bifurcation at the upper end (see Fig.~\ref{fig: t maze result} A). Each arm either contains a reward (+1) or a punishment (-7) while the other arm has the opposite. The bottom part of the maze contains a disambiguation cue that allows to distinguish whether the reward is in the right or in the left arm. Since the are only two possible environments (which we assume to have equal prior) and the cue and rewards are deterministic, the belief state is a single number in $\{0, 0.5, 1\}$. This allows us to organize the belief-augmented values into a $3 \times S$ table, where $S$ is the number of states (length of the maze). In this case, the dynamics of the belief state are very simple. The belief transitions from its starting $0.5$ to either $0$ or $1$ once either the cue or the reward/punishment locations are visited. The agent always starts in the middle of the maze, equally distant from the cue end and the reward/punishment ends. The agent can move up or down in the middle of the maze and down, left or right in the upper end. Each episode lasts $T = 20$ time steps. The agent collects reward/punishment at each time point as long as it is in the proper location in the maze. The optimal course of action is to reach the cue (information gathering) and then move to the other end to collect reward. In this problem there are $S \times 3$ possible belief-augmented states, where $S$ is the number of locations. As baseline, we use the classical (belief-augmented) tabular q-learning algorithm (Eq.~\ref{eq: belief-augmented update}) with $\epsilon$-greedy policy ($\epsilon = 0.5$) and decreasing learning rate $\eta_n = 0.01/(1 + 0.001 n)$, where $n$ is the epoch number. Our method is a tabular q-learning algorithm with PCR rewards. The value of future information is stored as a table of $S$ values since only the belief $0.5$ has non-zero value of future information. The cross-values are learned together with the value of future information from belief-augmented transitions without accessing the known ground truth environment using our belief horizon sampling approach. The value of future information is updated only after $100$ epochs so as to ensure some convergence of the cross q-values. Learning rates and $\epsilon$ were equal to the vanilla baseline. The results are visualized in Fig.~\ref{fig: t maze result}. These results are obtained using the greedy policy. The baseline algorithm quickly learns the correct deterministic dynamics but it does not learn the information gathering policy, leading to zero reward collection (see Fig.\ref{fig: t maze result} C). On the other hand, q-learning with PCR reward quickly learns the correct value function, leading to optimal reward collection in most repetitions (see Fig.\ref{fig: t maze result} B). 

\paragraph{The treasure map problem.}
\begin{figure}[ht]
    \centering
    \includegraphics[width=0.8\textwidth]{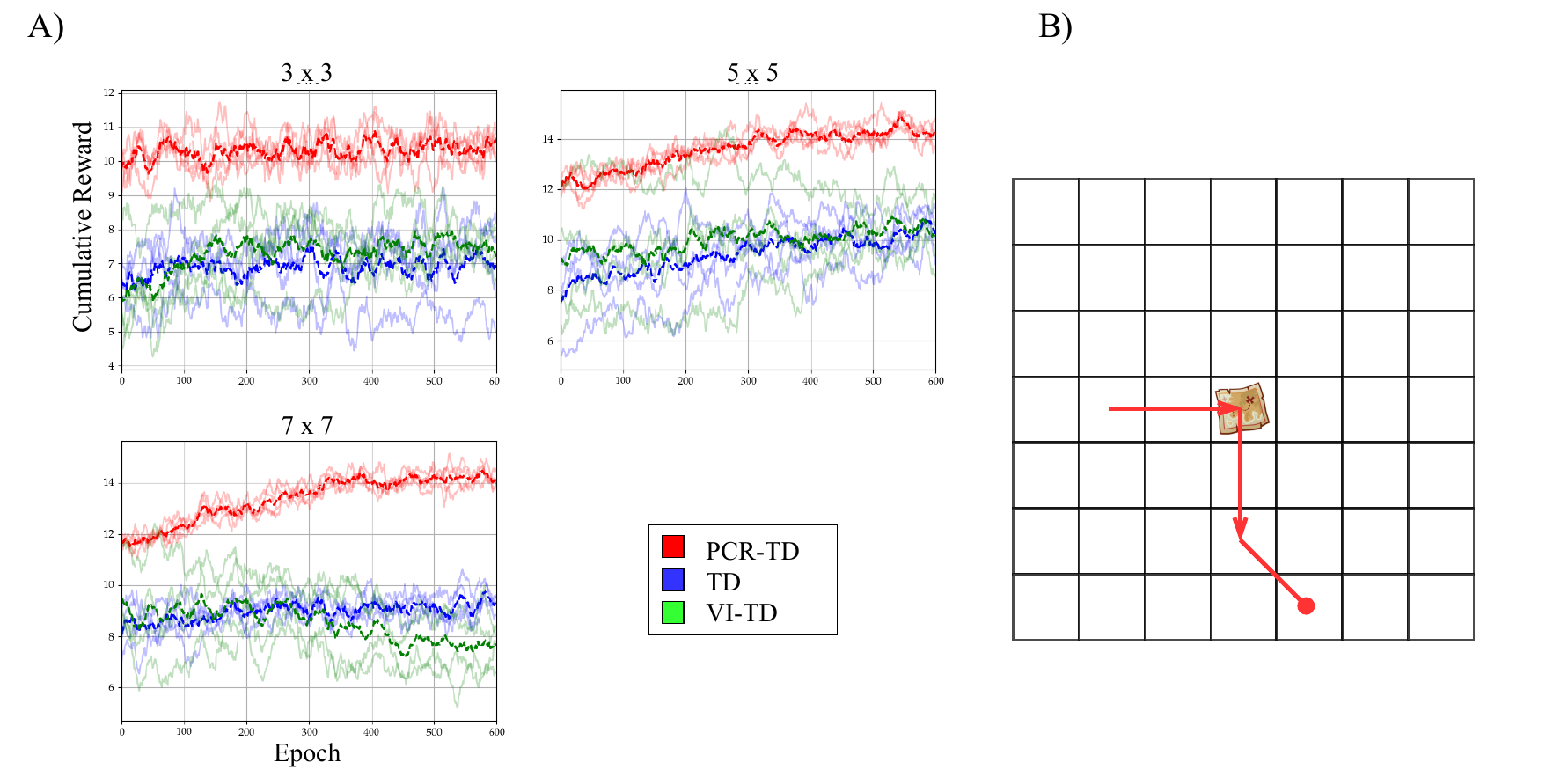}
   \caption{The "treasure map" experiment. A) Total training reward as function of the epoch of PCR-TD (red), TD (blue) and Vi-TD (green) for the three grid-world environments. The dashed line denotes the average over the runs. B) Example trajectory of trained PCR-TD. No other method learned how to systematically reach the map. }
    \label{fig: spatial}
\end{figure}
We now move to a more complex task where the agent needs to find a "treasure map" that reveals the reward distribution in a 2d environment. This is conceptually very similar to the "treasure mountain" problem used in \citep{zintgraf2021exploration}. We formalize this task as a spatial multi-armed bandit, where the rewards follow Bernoulli distributions and each cell in the grid-world has an independent reward probability. As in the conventional Bernoulli multi-armed bandit problem, the belief state can be updated analytically and is given by the counting parameters of the beta posterior distribution of each cell. When the agent visits a cell, reward is sampled from the appropriate Bernoulli distribution. Furthermore, the agent also observes 5 additional simulated "pulls" that update the belief state but do not give reward (this is an operationalization of the act of looking around and exploring the cell). Importantly, the central cell is a "treasure map" that allows the agent to observe 5 simulated "pulls" for each cell. In our proposed method (denoted as PRC-TD), the value of future information is given by the output of a convolutional network with two hidden layers that takes the spatial array of belief variables as input. Specifically, the network outputs the $w(x_t, b_t)$ function as in Eq.~\ref{eq: function approximation with bound} and the bound $\mathcal{B}(x_t, b_t)$ is estimated using $M=1$ and $N=40$. Instead of learning the cross-values with belief horizon sampling we compute them with (cross-)value iterations (see Appendix A). As main baseline, we use a more standard deep belief-augmented TD learning approach where the total belief-augmented value is given by the output of a deep network trained on the original reward structure. Since our method exploits the output of the value iterations, we also compared the results with a more sophisticated baseline (VI-TD) with the belief-augmented value given by the sum of the network output and the output of the value iterations applied to the expected reward probabilities (as inferred by the belief state). We also included VI-Greedy and VI-Thompson baselines, which implement greedy and Thompson sampling policies respectively on the values obtained by value iteration. All networks were trained for $2000$ epochs. In each epoch, a batch of $10$ environments (reward probabilities) were generated from the beta prior ($a = 0.1, b = 1$) and controlled sequences were simulated from the $\epsilon$-greedy policies. The TD loss was summed over all transitions and then backpropagated. Training performance was stored for each epoch. We selected the epoch with highest training performance for the test run. Test performance was quantified as (non-discounted) average total reward collected during $20$ trials on newly generated environments (averaged over a batch of $10$ environments per trial). Additional details about task, training and architectures are given in Appendix C. The total reward as function of the epoch can be seen in Fig.~\ref{fig: spatial} A for the $7 \times 7$ environments.  All results are given in Table \ref{tab: foraging} for the three environment sizes. Only our PCR-TD method was able to learn to visit the "treasure map" cell at the center in order to collect information and then switch to the optimal exploitation policy. A typical learned trajectory of this kind can be seen in Fig.~\ref{fig: spatial} B.

\begin{table}[]
\centering
\begin{tabular}{|l|l|l|l|l|l|}
\hline
                          & PRC-TD                 & TD            & VI-TD    & VI-Thompson & VI-Greedy \\ \hline
$3 \times 3$ &     $\boldsymbol{10.78 \pm 0.24}$                                           & $7.30 \pm 0.31$ &    $8.28 \pm 0.21$ & $10.41 \pm 0.16$ & $9.39 \pm 0.16$               \\ \hline
$5 \times 5$ & $\boldsymbol{15.17 \pm 0.22}$ & $10.46 \pm 0.37$  & $11.02 \pm 0.75$ & $11.72 \pm 0.12$    & $9.91 \pm 0.11$   \\ \hline
$7 \times 7$ & $\boldsymbol{15.33 \pm 0.16}$ & $10.42 \pm 0.26$ & $9.55 \pm 0.82$ & $10.41 \pm 0.15$    & $9.64 \pm 0.10$   \\ \hline
\end{tabular} \label{tab: foraging} \caption{Quantitative results of the treasure map experiment.}
\end{table}

\section{Conclusion \& Future work}
In the paper we showed that predictive reward cashing, with its associated policy that decouples the exploitation and exploration component, dramatically facilitates learning the Bayes-adaptive policy. While we developed the methods with a modular deep learning architecture in mind (see Fig.~\ref{fig: diagram}), in order to isolate the key contribution of the paper we limited our attention to settings where inference can be performed in closed-form. However, all approaches introduced here can in principle work together with approximate inference networks such as those used in \citep{zintgraf2019varibad} and \citep{zintgraf2021exploration}.

\appendix

\section{Computing cross-values by Bellman equation and value iterations}
In a finite horizon problem with known transition and reward distributions, the cross-values can be computed by backward iterations that are a straightforward extension of the bellman equation:
\begin{equation}\label{eq: bellman cross values}
 \left\{\begin{array}{@{}l@{}}
    v_t^{e_1}(x_{t}; e_1) = R \left(r_t \mid x_t, a_t, e_1 \right) + \gamma \sum_{x_{t+1}}  T\left(x_{t+1}\mid x_{t}, a_{t} , e_1\right) v_{t+1}^{e_1}(x_{t+1}; e_1) \\ 
    v_t^{e_1}(x_{t}; e_2) = R \left(r_t \mid x_t, a_t, e_2 \right) + \gamma \sum_{x_{t+1}}  T\left(x_{t+1}\mid x_{t}, a_{t} , e_2\right) v_{t+1}^{e_1}(x_{t+1}; e_2) \\ 
    a_t  = \underset{\alpha}{\text{argmax}} \left( R \left(r_t \mid x_t, a_t, e_1 \right) + \gamma \sum_{x_{t+1}}  T\left(x_{t+1}\mid x_{t}, a_{t} , e_1\right) v_{t+1}^{e_1}(x_{t+1}; e_1) \right)
  \end{array}\right.\,.
\end{equation}
This algorithm allows to efficiently compute the cross-values for an arbitrary pair of environments $e_1$ and $e_2$ in a highly parallelizable manner. As usual, the system of equations can be turned into value iterations in the infinite horizon setting simply by dropping the explicit time dependency from the value function.

\section{Convergence of belief horizon sampling}
Here we will prove that tabular TD learning training with belief horizon sampling convergences to the true cross-value table under the usual conditions of TD learning. Like the rest of this paper, this section will have a rather informal tone. However, all reasoning can be easily rigorously formalized.

We define the deterministic inference setting as a belief-augmented Markov decision process where the belief state can only transition from its initial prior state $b_0$ into one of the $S$ possible deterministic states, here denoted as $\delta_j$. These terminal states correspond to posterior distributions with zero entropy. In other words, each of them corresponds to a unique (non-augmented) Markov decision process. After a sequence of observed transitions $\{\ x_\tau, a_\tau, r_\tau\}_{\tau=0}^{t-1}$, the belief state $b_\tau$ can either be equal to $b_0$ or to one of the $\delta_j$s. In the latter case, at least one of the observed transitions has zero probability under all environments but one. On the other hand, in the former case all transitions have equal probability under all environments (this follows from the fact that the prior belief state has not been updated). 

In belief horizon sampling, the ground truth environment $e_1$ is sampled from the terminal belief state $b_T$. This environment is then used to select the proper entry of the cross-values table and to perform a TD update. The sampled environment $e_1$ can either be equal or different from the (inaccessible) true environment $e^*$ that generated the transitions. If it is equal then standard TD convergence results apply directly. Since all the $\delta_j$s represent deterministic posterior distributions, $e_1$ can be different from $e^*$ only when $b_T$ is equal to $b_0$. In this case, the sequence $\{\ x_\tau, a_\tau, r_\tau\}_{\tau=0}^{t-1}$ could be generated by any of the possible environments with equal probability and, consequently, the transitions can be used to update any of the cross-values tables as if they were sampled from their own environments without affecting convergence. 

\section{Details of the treasure map experiment}

\paragraph{Task details.}
In each episode, the agent starts in a random location in the grid-world and stays in the environment for $T = 25$ time steps. At each time step, the agent can move to each of the $8$ (except at the borders/corners) neighboring cells or stay in place. Reward is collected at each time step and is discounted with $\gamma = 0.96$.

\paragraph{Architecture details.}
The network output is a 2d array of value of future information, one for each spatial location. The network input was a tensor with two spatial dimensions and two channels, one for each parameter of the beta distributions. The network had two convolutional layers, the first with kernel sizes $ (3, 3)$ and $20$ output channels and the second with kernel sizes $(1, 1)$ and $1$ output channels. Relu activation functions were applied after the first convolutional layer. The output tensor was scaled by $0.01$ and then summed to a learnable constant table of values, one for each spatial location. 

\paragraph{Policy details.}
Predictively cashed reward $\tilde{\lambda}_t$ was computed from the cross-values using the approximation in Eq.~\ref{eq: approximate PCR} and $N = 80$ samples. We used an $\epsilon$-greedy training policy with respect to the value function. This does not require the training of a policy network since the transition model is known and deterministic. The TD loss for each transition was $\mathcal{L}(w) = ( \tilde{\lambda}_t + \gamma v(x_{t+1}, b_{t+1}; w) - v(x_{t}, b_{t}; w))^2$.


\paragraph{Baseline architecture details.}
The baseline TD algorithm has to learn a substantially more complex value function which, loosely speaking, includes both the exploitation and the exploration part of the value. For these reasons, we thought it more effective to use a fully connected architecture as it is in theory capable of learning arbitrarily complex dependencies between the input variables. We therefore used a two-layers fully connected architecture and $4 H^2$ hidden layers, where $H$ is the linear size of the grid-world. To facilitate learning, the output of the network corresponding to the $j,k-th$ grid point was $\rho_{jk} + f_{jk}(b_t)$, where $f_{jk}(b_t)$ denotes the output of the network and $\rho_{jk}$ is the expected reward probability of the cell given the current belief. This significantly improves performance. Since our method uses the solution of the value iterations, we also used an alternative method (VI-TD) with the belief-augmented value given by $v(\rho_{jk}) + f_{jk}(b_t)$, with $v(\rho_{jk})$ being the value obtained from the expected values by value iteration.

\bibliographystyle{iclr2022_conference.bst}
\bibliography{ref.bib}
\end{document}

%% file: math_commands.tex
\usepackage{amsmath,amsfonts,bm}

\def\eqref#1{equation~\ref{#1}}

\def\1{\bm{1}}

\DeclareMathAlphabet{\mathsfit}{\encodingdefault}{\sfdefault}{m}{sl}
\SetMathAlphabet{\mathsfit}{bold}{\encodingdefault}{\sfdefault}{bx}{n}

